\documentclass[letterpaper]{article} 
\usepackage{aaai2026}  
\usepackage{times}  
\usepackage{helvet}  
\usepackage{courier}  
\usepackage[hyphens]{url}  
\usepackage{graphicx} 
\usepackage{natbib}  
\usepackage{caption} 
\usepackage{algorithm}
\usepackage{algorithmic}

\usepackage{newfloat}
\usepackage{listings}
\DeclareCaptionStyle{ruled}{labelfont=normalfont,labelsep=colon,strut=off} 
\floatstyle{ruled}
\newfloat{listing}{tb}{lst}{}
\floatname{listing}{Listing}
\usepackage{amsmath}
\usepackage{cleveref}
\usepackage{subfig}
\usepackage{amsfonts}
\usepackage{algorithm}
\usepackage{algorithmic}
\usepackage{amssymb}
\usepackage{graphicx}
\usepackage{multirow}
\usepackage{chngcntr}
\usepackage{adjustbox}
\usepackage{booktabs}
\usepackage{amssymb}

\title{OTARo: Once Tuning for All Precisions toward Robust On-Device LLMs}
\author {
    Shaoyuan Chen\textsuperscript{\rm 1,\rm 2,$\dagger$,$\triangle$},
    Zhixuan Chen\textsuperscript{\rm 1,$\dagger$},
    Dawei Yang\textsuperscript{\rm 1,$\ast$,$\dagger$},
    Zhihang Yuan\textsuperscript{\rm 3},
    Qiang Wu\textsuperscript{\rm 1}
}
\affiliations {
    \textsuperscript{\rm 1}Houmo AI\\
    \textsuperscript{\rm 2}Sun Yat-sen University\\
    \textsuperscript{\rm 3}Peking University\\
    chenshy299@mail2.sysu.edu.cn, zhixuan.chen@houmo.ai, dawei.yang@houmo.ai, yuanzhihang@pku.edu.cn, qiang.wu@houmo.ai
}

\usepackage{bibentry}

\begin{document}

\maketitle

\def\thefootnote{$\ast$}\footnotetext{Corresponding author}
\def\thefootnote{$\dagger$}\footnotetext{Equal contribution}
\def\thefootnote{$\triangle$}\footnotetext{Work done as an intern at Houmo AI}

\begin{abstract}
Large Language Models (LLMs) fine-tuning techniques not only improve the adaptability to diverse downstream tasks, but also mitigate adverse effects of model quantization. Despite this, conventional quantization suffers from its structural limitation that hinders flexibility during the fine-tuning and deployment stages. Practical on-device tasks demand different quantization precisions (i.e. different bit-widths), e.g., understanding tasks tend to exhibit higher tolerance to reduced precision compared to generation tasks. Conventional quantization, typically relying on scaling factors that are incompatible across bit-widths, fails to support the on-device switching of precisions when confronted with complex real-world scenarios. To overcome the dilemma, we propose OTARo, a novel method that enables on-device LLMs to flexibly switch quantization precisions while maintaining performance robustness through once fine-tuning. OTARo introduces Shared Exponent Floating Point (SEFP), a distinct quantization mechanism, to produce different bit-widths through simple mantissa truncations of a single model. Moreover, to achieve bit-width robustness in downstream applications, OTARo performs a learning process toward losses induced by different bit-widths. The method involves two critical strategies: (1) Exploitation-Exploration Bit-Width Path Search (BPS), which iteratively updates the search path via a designed scoring mechanism; (2) Low-Precision Asynchronous Accumulation (LAA), which performs asynchronous gradient accumulations and delayed updates under low bit-widths. Experiments on popular LLMs, e.g., LLaMA3.2-1B, LLaMA3-8B, demonstrate that OTARo achieves consistently strong and robust performance for all precisions.
\end{abstract}

\section{Introduction}
\begin{figure}
    \centering
    \includegraphics[width=1\linewidth, trim={0pt 0pt 450pt 345pt}, clip]{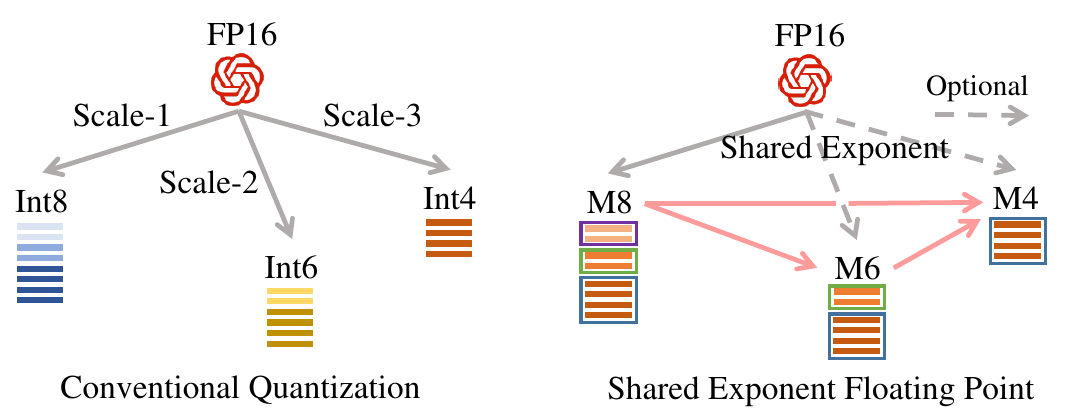}
    \caption{A comparison of conventional quantization and Shared Exponent Floating Point (SEFP) in supporting dynamic precision switching. Gray arrows represent quantization, while red arrows indicate cross-precision conversion. M8, M6, and M4 denote mantissa bit-widths in SEFP.}
    \label{sefp_vs_ssfp}
\end{figure}

In recent years, Large Language Models (LLMs) have shown outstanding capabilities in Natural Language Processing (NLP) \cite{achiam2023gpt,zubiaga2024natural}, and on-device deployment of LLMs has emerged as a frontier research direction to enable real-time responsiveness, privacy, and personalization in intelligent terminal services \cite{qu2025mobile,tang2025scaling}. With advances in processing power, memory bandwidth, and storage capacity of edge devices, on-device deployment of compact LLMs becomes feasible.
In parallel, model compression techniques, e.g., quantization \cite{achiam2023gpt,lin2024awq}, 
reduce deployment costs while incurring some accuracy degradation.

For a better inference performance of LLMs in devices, fine-tuning techniques play a pivotal role by not only enhancing the adaptability across downstream tasks, but also effectively mitigating the accuracy degradation induced by quantization \cite{yang2024unveiling,lu2025fine}. Despite this, conventional quantization still faces challenges in complex real-world scenarios due to its inherent limitation in switching bit-widths.

Practical on-device tasks require different quantization precisions, corresponding to different bit-widths. For instance, generation tasks tolerate increased inference time in exchange for a higher precision, whereas understanding tasks can obtain immediate responses with a lower precision \cite{manduchi2024challenges}. In addition, different precisions can be employed in the prefill and decoding phases to optimize the inference efficiency \cite{qiao2025tellme}. Consequently, the practical on-device deployment necessitates support for all precisions, rather than a model with a fixed precision. 

In conventional quantization, the original weights are scaled and clipped into different numerical ranges depending on the bit-width, which prevents the reuse of scaling factors (scales) across precisions. Once quantized to a specific bit-width, the model loses the flexibility to switch between precisions. 
Edge devices can maintain a model zoo containing models of each precision, which substantially increases storage overhead and extends the total fine-tuning time. 
Alternatively, compressing a full- or half-precision model into various quantization precisions at runtime still requires offline fine-tuning for bit-specific and high-quality scaling factors, thereby incurring similar fine-tuning burdens. It also increases storage demands and introduces non-negligible computational costs during on-device quantization.

To address the challenges, this paper proposes OTARo (\textbf{O}nce \textbf{T}uning for \textbf{A}ll Precisions toward \textbf{Ro}bust On-Device LLMs), which obtains one unified model through once fine-tuning to support all quantization precisions. OTARo introduces Shared Exponent Floating Point (SEFP), a distinct quantization mechanism. Notably, compared to conventional quantization, SEFP eliminates scaling factors and allows flexible switching of precisions, as illustrated in \cref{sefp_vs_ssfp}. In SEFP, an exponent is shared in each group of parameters, and individual mantissas are maintained for each parameter. After aligning exponents, any expected bit-width can be achieved through simple mantissa truncation, and the higher bit-widths can be easily converted to the lower ones.

Furthermore, to enhance performance robustness across bit-widths, OTARo introduces an efficient learning scheme that jointly optimizes for mixed quantization losses induced by different SEFP configurations. The learning process incorporates an iterative bit-width selection, and adopts delayed updates to mitigate the effects of low-precision training.
Generally, by unifying all precisions within one fine-tuned model and obviating the need for separate fine-tuning per bit-width, 
OTARo substantially enhances resource efficiency and deployment flexibility in practical applications.

The contributions of our work are summarized as follows:
\begin{itemize}
    \item We propose OTARo, a novel fine-tuning paradigm for building robust on-device LLMs. 
    OTARo produces a unified model through once fine-tuning to support all quantization precisions. 
    It introduces Shared Exponent Floating Point (SEFP) to enable lightweight and hardware-friendly multi-precision adaptation, and jointly update weights for quantization errors from different bit-widths.
    \item We identify the commonality on gradient directions under all bit-width settings, and then propose Exploitation-Exploration Bit-Width Path Search (BPS) strategy to determine the optimal sequence of bit-widths.
    \item We observe that low-precision training induces severe gradient oscillations. To mitigate the oscillations, we conduct high-dimensional projection analysis on the gradients, and propose Low-Precision Asynchronous Accumulation (LAA) strategy, which performs asynchronous gradient accumulations and delayed updates.
    \item 
    We conduct the comprehensive evaluations on popular LLMs, e.g., LLaMA3.2-1B and LLaMA3-8B. Experimental results show that OTARo consistently outperforms baselines across all tested bit-widths, in both task-specific and zero-shot settings, highlighting its robustness for multi-precision deployment of LLMs.

\end{itemize}

\section{Related Work}

\subsection{Quantization}
Quantization is a critical technique to reduce the memory footprint and inference costs \cite{wei2024advances}. 
Post-Training Quantization (PTQ) methods quantize a pre-trained model without retraining. GPTQ \cite{frantar2022gptq} uses arbitrary order quantization, lazy batch updates, and Cholesky reformulation.
CFWS \cite{yang2024post} introduces coarse \& fine weight splitting and an enhanced KL calibration metric.
AWQ \cite{lin2024awq} finds the salient weights based on the activation distribution, and scales the salient weight channels through an equivalent transformation.
QuaRot \cite{ashkboos2024quarot} uses Hadamard transform and computational invariance to mitigate outliers in activations. 
SpinQuant \cite{liu2024spinquant} uses a learnable rotation matrix that is fused into weights.
Any-Precision LLM \cite{park2024any} performs incremental upscaling and specialized hardware design to merge multiple integer formats.
QUARK \cite{zhao2025quark} proposes a reordering-based
group quantization scheme.
Quantization-Aware Training (QAT) methods incorporate fake quantization in training, allowing models to lea quantization noises \cite{liu2023llm,chen2024efficientqat}. 

\subsection{Shared Exponent Floating Point}
\begin{figure*}[!t]
    \centering
    \includegraphics[width=1
    \linewidth, trim={0pt 0pt 0pt 400pt}, 
                     clip]{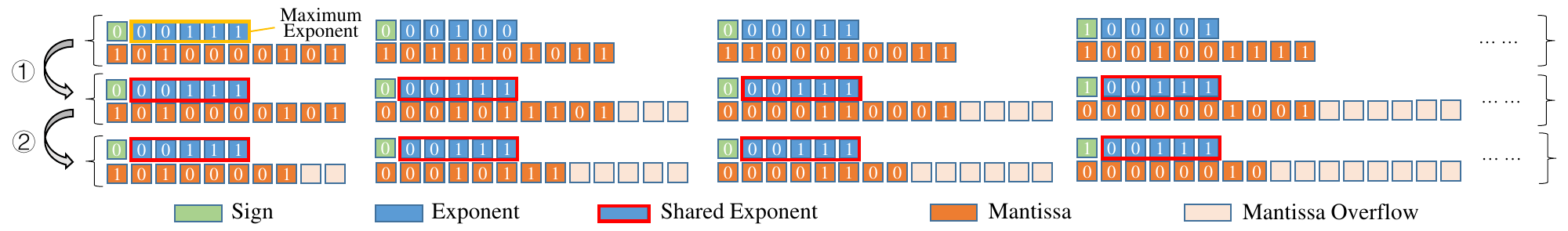}
    \caption{A simple illustration of SEFP quantization (e.g., from FP16 to E5M8) of a group of parameters. Step 1 shows the exponent alignment and mantissa right-shifting for FP16 values. Step 2 shows the forced mantissa truncation.} 
    \label{fig-sefp}
\end{figure*}

Shared Exponent Floating Point (SEFP) quantization shares an exponent in each parameter group while maintaining individual mantissas for each parameter \cite{gao2024precision,zhou2025gsq}.
SEFP quantization procedure consists of two main steps, illustrated in \cref{fig-sefp}. 
Firstly, the maximum exponent of each parameter group is selected as the shared one before each mantissa is shifted right according to the difference between the shared exponent and its original one. Secondly, shifted mantissas are truncated to a fixed width. 
SEFP bit-width is typically donated as E$e$M$m$, where $e$ and $m$ denote the bit-width of exponents and mantissas, respectively.
Accuracy declines arise almost from the forced truncation in Step 2; the impacts of mantissa overflow in step 1 is negligible.
\cite{kalliojarvi2002roundoff}. 

\section{Method}
\subsection{Preliminaries}
Due to the non-differentiable operations involved in SEFP quantization, such as mantissa right-shifting and truncation, we incorporate the Straight-Through Estimator (STE) \cite{yin2019understanding} to approximate the gradients, 
and introduce SEFP quantization errors into training losses to facilitate the learning process.
The approach is formulated as follows:

\begin{equation}
       \frac{\partial Q(\omega,b)}{\partial \omega} = 1 
\end{equation}

\begin{equation}
    \mathcal{L} = \text{Loss}(y,Q(\omega,b)x)
\end{equation}

\begin{equation}
    \frac{\partial \mathcal{L}}{\partial \omega}= \frac{\partial \mathcal{L}}{\partial Q(\omega,b)} \frac{\partial Q(\omega,b)}{\partial \omega}
    = \frac{\partial \mathcal{L}}{\partial Q(\omega,b)}
\end{equation}
where, $\omega$ is weight, $Q(\cdot,b)$ is SEFP function targeting bit-width $b$, $x$ is input, $y$ is label, $\text{Loss}(\cdot)$ is training loss function. In the above manner, OTARo enables models to adapt to quantization errors from different SEFP bit-widths, and fixed precision fine-tuning only learns for quantization errors from a specific bit-width.

\subsection{Problem Statement}
In OTARo, once fine-tuning is performed to obtain a model robust to all bit-widths in SEFP. Each bit-width corresponds to a certain level of precision, with higher bit-widths yielding higher precision. The objective can be formulated as:

\begin{equation}
     \omega = \arg\min_{\omega} \frac{1}{|\mathcal{B}|}\sum_{b\in\mathcal{B}}\mathbb{L}(Q(\omega,b))
\end{equation}
where, $b$ is the bit-width,  $\mathcal{B}$ is the set of bit-widths, $\mathbb{L}(\cdot)$ is test loss function, and $Q(\omega,b)$ is SEFP function under bit-width $b$. 
We aim to develop a single model to maintain high performance across all bit-widths, achieving accuracy comparable to the original model before quantization, thus supporting robust and adaptable on-device deployment of LLMs.

\subsection{Exploitation-Exploration Bit-Width Path Search}

\begin{figure}
    \centering
    \includegraphics[width=1\linewidth, trim={0pt 25pt 0pt 0pt}]{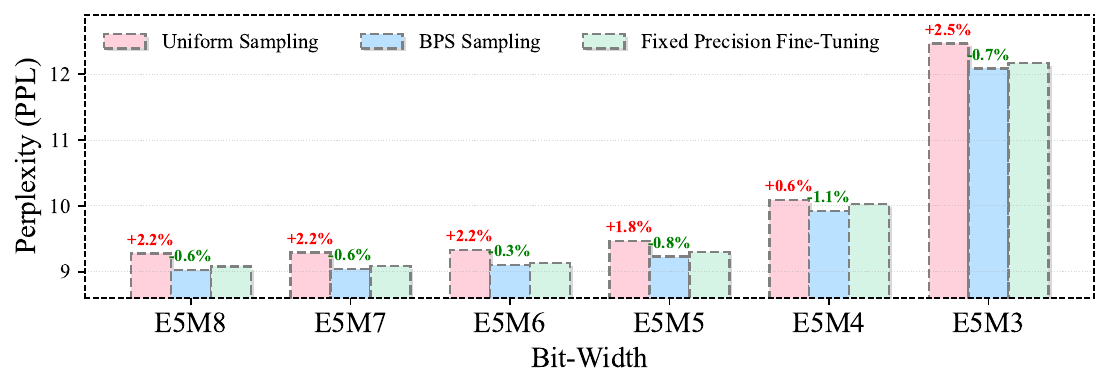}
    \caption{A comparison of uniform sampling, BPS sampling and fixed precision fine-tuning. We report perplexity changes of the first two approaches relative to the last one. In the experiments, LLaMA3.2-1B is fine-tuned and evaluated on the WikiText2 train and test set, respectively.}
    \label{fig-sample}
\end{figure}

\begin{figure*}[!t]
    \centering
    \subfloat[q proj]{
    		\includegraphics[scale=0.36, trim={10pt 0pt 0pt 10pt}, 
                     clip]{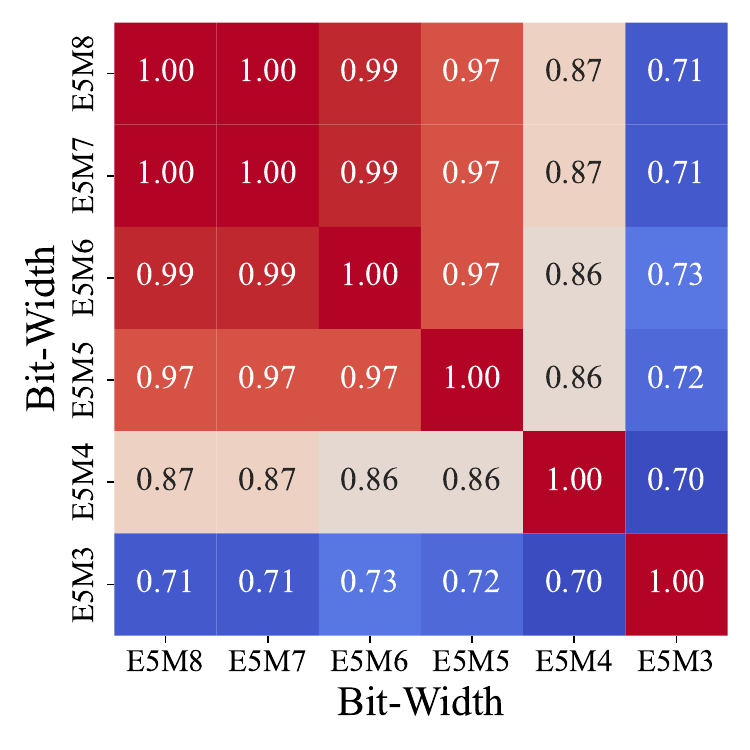}}
    \subfloat[k proj]{
    		\includegraphics[scale=0.36, trim={0pt 0pt 0pt 10pt}, 
                     clip]{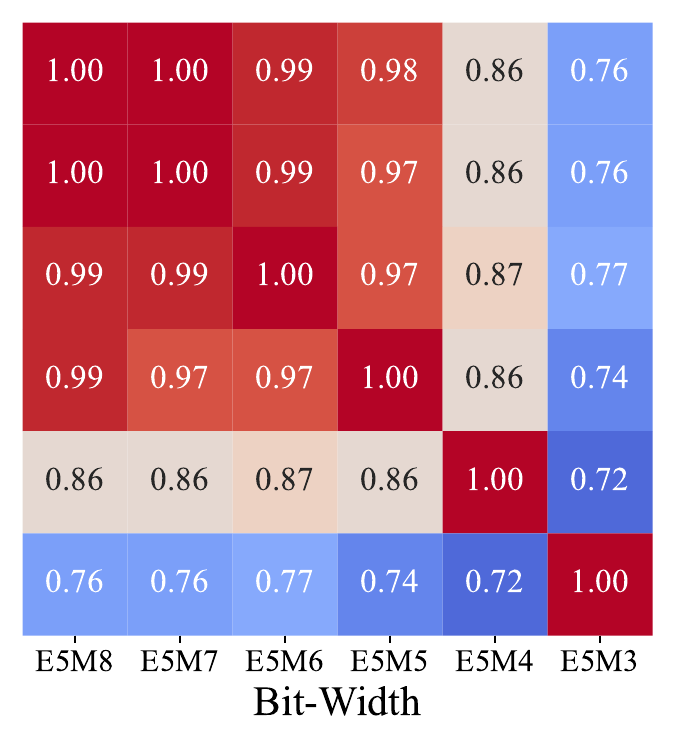}}
    \subfloat[v proj]{
    		\includegraphics[scale=0.36, trim={0pt 0pt 0pt 10pt}, 
                     clip]{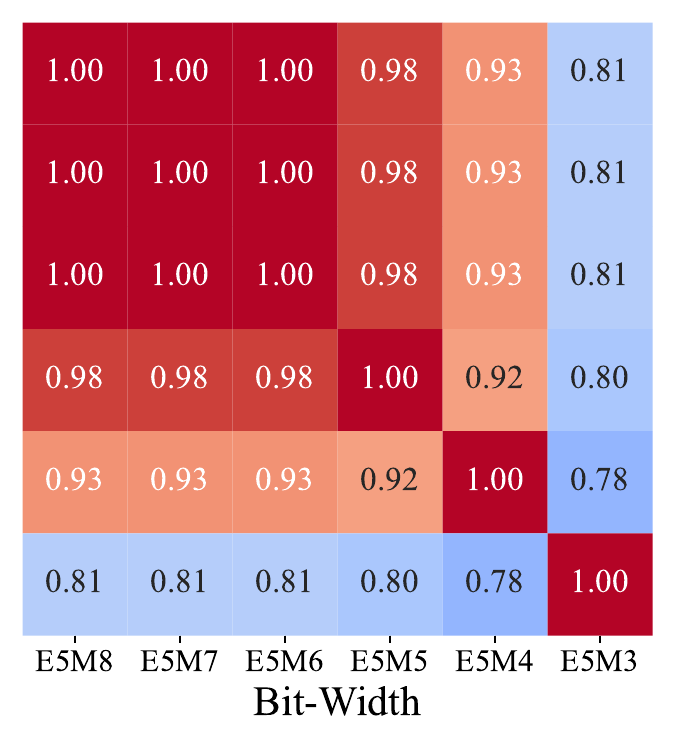}}
    \subfloat[down proj]{
    		\includegraphics[scale=0.365, trim={0pt 0pt 0pt 5pt}, 
                     clip]{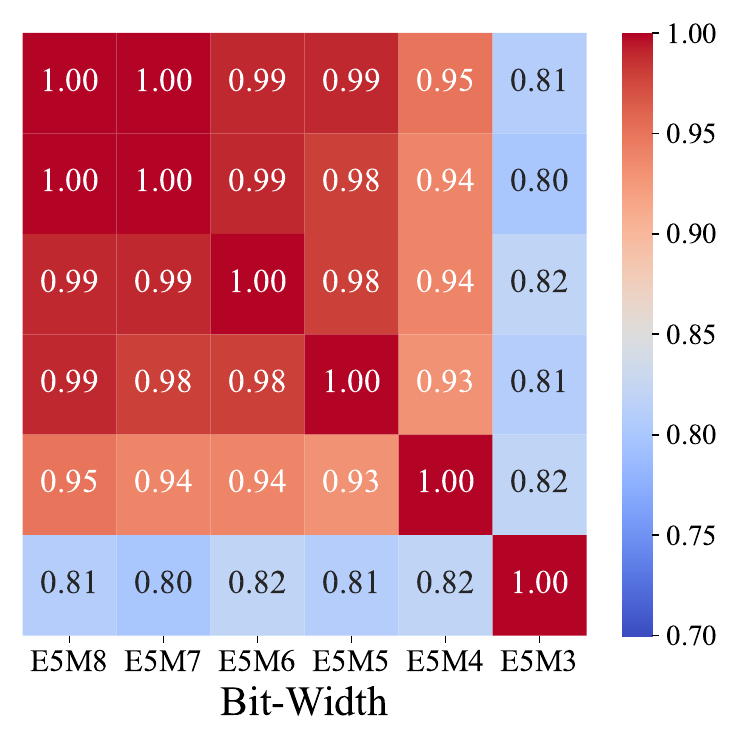}}
    \caption{Cosine similarities between the gradients produced by different bit-widths of the LLaMA3.2-1B layer-15 q/k/v/down projector. There exists a certain degree of similarity between gradients under different bit-widths. Furthermore, the gradient at each bit-width tends to exhibit stronger similarity with that of its higher bit-width. For example, for q projector, the cosine similarity of gradients between E5M5 and E5M8/E5M7/E5M6 is 0.97, whereas the cosine similarity with E5M4 and E5M3 decreases to 0.86 and 0.72, respectively.}
    \label{fig-cos}
\end{figure*}

To ensure robustness across bit-widths, an intuition is to uniformly sample all target bit-widths, and facilitate models to update based on the gradients induced by quantization errors of different bit-widths. 
However, our empirical results indicate that this straightforward sampling approach does not align with fixed precision fine-tuning in performance, as shown in \cref{fig-sample}. This finding prompts us to rethink: whether different bit-widths should be sampled non-uniformly, and scheduled in a deliberate order during fine-tuning. 

To assess the feasibility of sampling all bit-widths for robustness, and to inform the design of a practical implementation strategy, we revisit bit-width sampling from a gradient-centric perspective, as illustrated in \cref{fig-cos}.

It can be observed that the gradients produced at different bit-widths exhibit a certain degree of similarity, which supports the feasibility of enhancing robustness by sampling all bit-widths. Furthermore, for both the higher and lower bit-widths, the gradient direction shows higher consistency with those of the higher ones, but less consistency even with that of the adjacent lower one. This commonality suggests the existence of a larger shared potential subspace between higher bit-widths and others in terms of gradient directions. 

Motivated by the observation and analysis, we argue that the bit-width sampling should not only explore all bit-widths to sufficiently learn the loss landscapes across them, but also gradually update toward higher bit-widths that exhibit stronger alignment with the others in gradient directions.

Accordingly, we propose the Exploitation-Exploration Bit-Width Path Search (BPS) strategy, which iteratively selects the bit-width that achieves the highest score in a designed scoring mechanism. The scoring mechanism is formulated as follows:
\begin{equation}
        \text{Score}(b) = \lambda\sqrt{\frac{\ln t}{t_b}}-\mathcal{L}_b
\end{equation}
where, the score of bit-width $b$ is denoted as $\text{Score}(b)$, $\mathcal{L}_b$ is the real-time loss for bit-width $b$, $t$ is the current number of training batches, $t_b$ is the search frequency of $b$, and exploration coefficient $\lambda$ is a constant multiplied by the exploration term. 

This design encourages the bit-width search path to explore underutilized bit-widths. As fine-tuning progresses, the exploration term of frequently used bit-widths gradually decreases, naturally leading to sampling underutilized ones. The dynamic balance ensures diversity in bit-width selection. 
This design also ensures that, as the number of batches increases, higher bit-widths are selected more frequently than lower ones. 
To validate this convergence property, we examine whether the score discrepancy between the higher and lower bit-widths consistently converges toward a positive value as $t$ increases.
Suppose $h$ and $l$ denote two distinct bit-widths, and $h$ is the higher one. The score difference is defined as:
\begin{equation}
    \Delta =(\lambda \sqrt\frac{\ln t}{t_h}-\mathcal{L}_h)-(\lambda \sqrt\frac{\ln t}{t_l}-\mathcal{L}_l)
\end{equation}
where, $T$ is the total number of batches, $t_h$, $t_l$ are respectively the current counts of $h$ and $l$ being searched (approximately increasing linearly), $\mathcal{L}_h$, $\mathcal{L}_l$ are the corresponding losses at bit-widths $h$ and $l$, with $\mathcal{L}_l$ generally being larger due to the lower precision of $l$. We reorganize the terms in the above equation:
\begin{equation}
        \Delta=(\mathcal{L}_{l}-\mathcal{L}_h)+\lambda (\sqrt{\frac{t}{t_h} }- \sqrt{\frac{t}{t_l}} )\sqrt{\frac{\ln t}{t}}
\end{equation}

As $t$ increases to a high value, $\Delta$ approaches $\mathcal{L}_l - \mathcal{L}_h$, which is increasingly likely to be positive:
\begin{equation}
    \lim_{t\rightarrow T}\sqrt{\frac{\ln t}{t}} \rightarrow 0
\end{equation}
\begin{equation}
    \lim_{t\rightarrow T}\Delta \rightarrow \mathcal{L}_{l}-\mathcal{L}_h
\end{equation}

Therefore, the search path in the BPS strategy gradually converges toward the higher bit-widths with smaller losses and more robust gradient directions. As show in \cref{fig-sample}, the sampling based on the BPS strategy can match or even outperform fixed-precision fine-tuning in each bit-width.

\subsection{Low-Precision Asynchronous Accumulation}
\begin{figure}[!t]
    \centering
    \includegraphics[width=0.96\linewidth, trim={0pt 0pt 20pt 40pt}, 
                     clip]{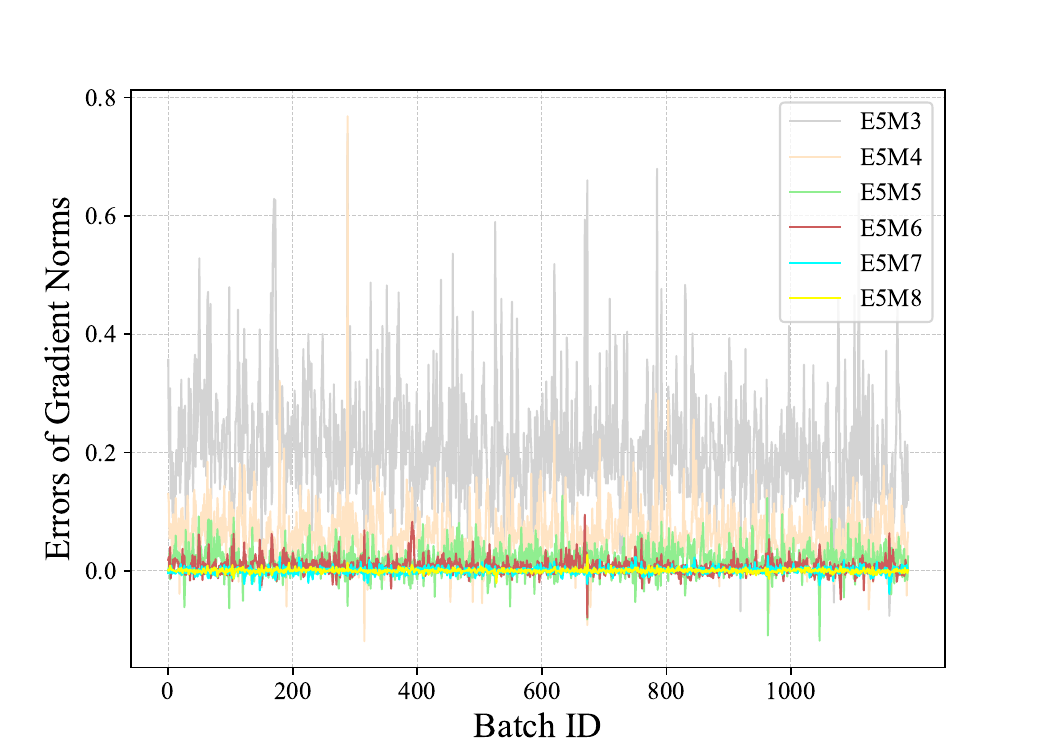}
    \caption{The errors of gradient norms $||\nabla_{\text{sefp}}||-||\nabla_{\text{fp}}||$ under different SEFP bit-widths. Gradients are calculated on LLaMA3.2-1B layer-15 down projector.}
    \label{fig-norms}
\end{figure}

Through the BPS strategy, we obtain a search path that sufficiently explores multiple bit-widths while gradually favoring higher ones. However, the robustness of fine-tuned models suffers from the impacts of low bit-widths. To address this issue, we propose the Low-Precision Asynchronous Accumulation (LAA) strategy, which utilizes a distinctive weight update scheme under continuous low-bit width sampling conditions. To demonstrate the problem, we perform a detailed analysis of errors introduced by SEFP quantization:
\begin{align}
    \mathcal{L}_q&=\frac{1}{2}||xQ(\omega,m)-x\omega||^2 
    \\
    &= \frac{1}{2}||\frac{x[\omega 2^m]}{2^m}-x\omega||^2 
\end{align}

\begin{equation}
        \frac{\partial \mathcal{L}_q}{\partial \omega} = \frac{x \left( \omega 2^m - [\omega 2^m] \right)}{2^m} \cdot \frac{\partial Q(\omega,m)}{\partial \omega}
\end{equation}
where, $\mathcal{L}_q$ denotes an approximation of the SEFP error, $\partial \mathcal{L}_q / \partial \omega$ is the derivative of the error with respect to the weight., $m$ denotes the number of mantissa bits in SEFP, and $[ \cdot ]$ is rounding function. Suppose $\omega_0$ is an arbitrary parameter in $\omega$, we define the function $\epsilon(\omega_0)$ as:
\begin{equation}
    \epsilon(\omega_0)=\frac{\omega_0 2^m - [\omega_0 2^m]}{2^m}
\end{equation}

Function $\epsilon(\omega_0)$ describes a sawtooth wave with both period and amplitude equal to $1/2^m$ (see Appendix A for visualization). Due to the discontinuities introduced by the rounding operation, the function linearly increases from 0 within each period, then abruptly drops to a negative value before jumping back to 0, forming a periodic oscillatory pattern. 
When $m$ is small, the wave exhibits larger amplitudes, leading to more pronounced oscillations.
In low bit-width settings, changes in parameters can induce significant and periodic variations in SEFP quantization errors, resulting in periodic and intense oscillations in both loss and gradient during fine-tuning. 

\begin{figure}
    \centering
    \includegraphics[width=0.8\linewidth, trim={160pt 20pt 100pt 70pt}]{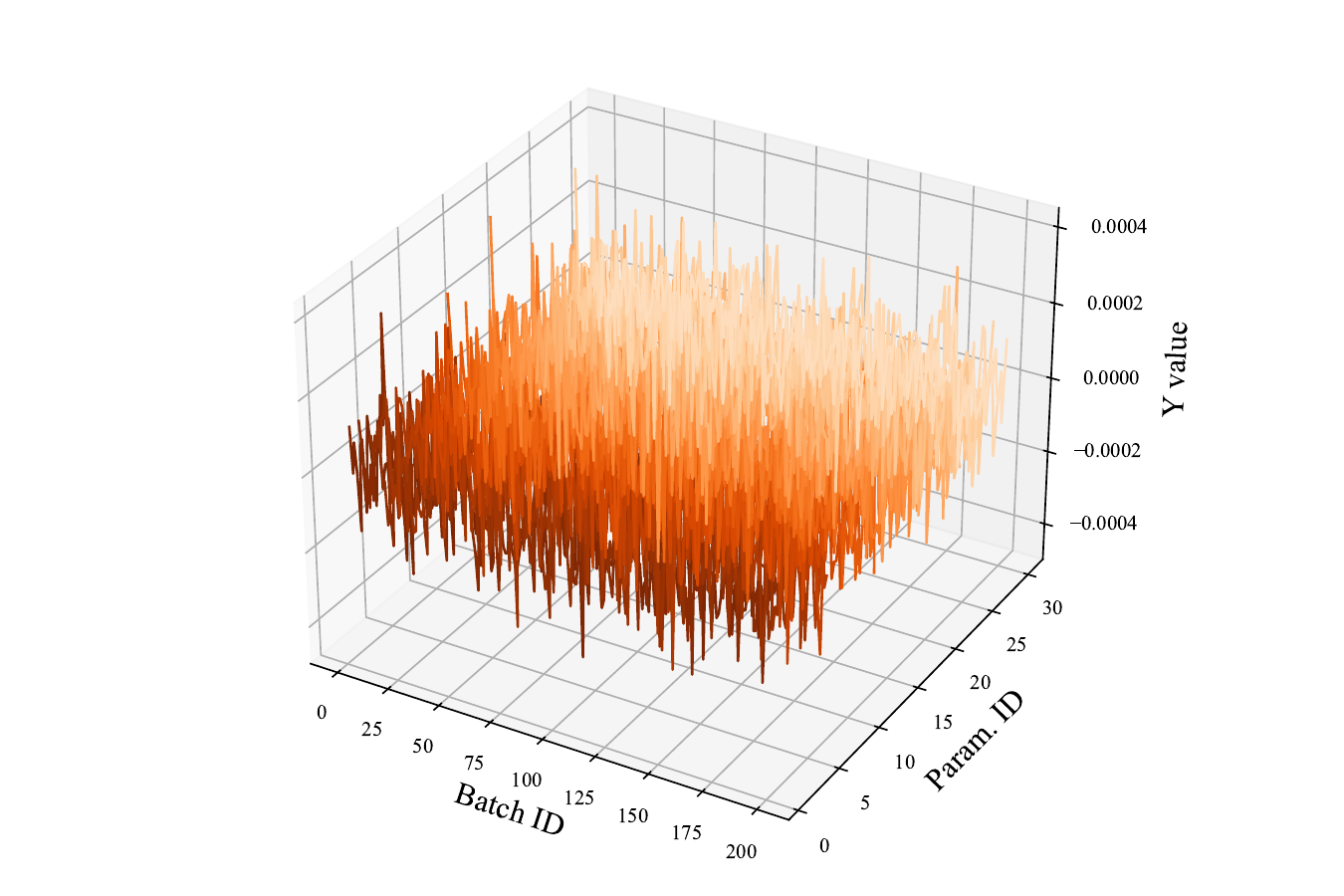}
    \caption{$Y$ values. For model LLaMA3.2-1B and dataset WikiText2, the figure shows the $Y$ values of 30 gradient values in 200 batches, and indicates that $\mathbb{E}[Y] \approx 0$.}
    \label{fig-gradhills}
\end{figure}
Suppose that the gradient in FP16, SEFP fine-tuning are respectively $\nabla_{\text{fp}}$ and $\nabla_{\text{sefp}}$, the error of gradient norms induced by SEFP is $||\nabla_{\text{sefp}}||-||\nabla_{\text{fp}}||$. As shown in \cref{fig-norms}, we observe that the oscillations in $||\nabla_{\text{sefp}}||-||\nabla_{\text{fp}}||$ become more intense as the bit-width decreases, and the oscillation exhibits an overall periodic pattern, providing indirect support for the above inference. 

\begin{algorithm}[t]
\caption{Overall Pipeline of OTARo.}
\label{alg}
\begin{algorithmic}[1]

\REQUIRE Batch id $t$, total number of batches $T$, bit-width $b$, training loss function $\text{Loss}()$, weight $\omega$, quantized $\omega$ under a bit-width $b$ $Q(\omega,b)$, exploration coefficient $\lambda$, search count of a bit-width $b$ $t_b$, inputs $x$, labels $y$, batch id and also a flag for gradient accumulations $i$ (starts from 0), delayed interval $N$, learning rate $\eta$.

\FOR{each batch $t = 1$ to $T$}

    \STATE Calculate scores: 
    $
    \text{Score}(b) = \lambda \sqrt{(\ln t) / t_b} - \mathcal{L}_b
    $
    \STATE Select optimal bit-width $b^* = \arg\max_{b} \text{Score}(b)$
    \STATE Calculate the training loss: $\mathcal{L} = \text{Loss}(y,Q(\omega,b^*)x)$
    \STATE Calculate the gradient: $\nabla_{\text{sefp}} = \partial \mathcal{L} / \partial Q(\omega,b)$
    
    \IF{bit-width is ultra-low}
        \IF {$i$ is $0$}
            \STATE Initialize the general gradient: $\nabla_{\text{sefp}}^N \leftarrow \nabla_{\text{sefp}}$
        \ELSE
            \STATE Accumulate gradients:
            $
            \nabla_{\text{sefp}}^N \leftarrow \nabla_{\text{sefp}}^N + \nabla_{\text{sefp}}
            $
        \ENDIF
        \STATE $i \leftarrow i + 1$
        \IF{$i$ is $N$}
            \STATE Update weights:
            $
            \omega \leftarrow \omega - \eta \nabla_{\text{sefp}}^N
            $
            \STATE $i \leftarrow 0$
        \ENDIF
    \ELSE
        \STATE Standard gradient updates:
        $
        \omega \leftarrow \omega - \eta \nabla_{\text{sefp}}
        $
    \ENDIF

\ENDFOR

\end{algorithmic}
\end{algorithm}

Based on these analysis, the error introduced by SEFP quantization is a critical issue to be tackled under low bit-widths. 
To measure the space distance between gradients with and without introducing SEFP quantization errors (i.e., \(\nabla_{\text{fp}}\) and \(\nabla_{\text{sefp}}\), respectively), we model the relationship of them through a high-dimensional linear mapping:
\begin{equation}
    \nabla_{\text{sefp}} = X \cdot \nabla_{\text{fp}} + Y
\end{equation}
where $X$ denotes the global linear mapping matrix from $\nabla_{\text{fp}}$ to $\nabla_{\text{sefp}}$, and $Y$ represents the residual perturbation term introduced by quantization, which corresponds to the non-linear and unexplained quantization noise.
Estimation of \(X\),\(Y\) is described in Appendix B, using Least Squares Method (LSM).

To empirically validate the statistical properties of the perturbation \(Y\), we conduct experiments under the low bit-width setting (e.g., E5M3). As shown in \cref{fig-gradhills}, the results indicate that while \(Y\) exhibits considerable fluctuation between batches, its mean remains close to zero, that is,
\begin{equation}
    \mathbb{E}[Y] \approx 0
\end{equation}

In the LAA strategy, gradients are asynchronously accumulated over \(N\) batches to generate a gradient $\nabla_{\text{sefp}}^N$ to update models. $\nabla_{\text{sefp}}^N$ can be expressed as:

\begin{equation}
    \nabla_{\text{sefp}}^N =
    \sum_{i=1}^N \nabla_{\text{sefp},i} = X \sum_{i=1}^N \nabla_{\text{fp},i} + \sum_{i=1}^N Y_i.
\end{equation}

Suppose independence or weak correlation among the \(Y_i\), the relative influence of the perturbation with $N$ rising diminishes as:

\begin{equation}
    \frac{\left\|\sum_{i=1}^N Y_i\right\|}{\left\|\sum_{i=1}^N \nabla_{\text{fp},i}\right\|} \propto \frac{1}{\sqrt{N}} \to 0
\end{equation}

Accordingly, the parameter update rule under the LAA strategy becomes:

\begin{equation}
    \omega \leftarrow \omega - \eta \sum_{i=1}^N \nabla_{\text{sefp},i} = \omega - \eta X \sum_{i=1}^N \nabla_{\text{fp},i} - \eta \sum_{i=1}^N Y_i
\end{equation}
where \(\eta\) is the learning rate. For the cumulative perturbation tends toward zero expectation, the high-frequency oscillations introduced by low bit-widths are effectively mitigated. This stabilizes the training process and enhances convergence. In addition, asynchronous accumulation avoids the memory growth in the computing device.

Finally, the pseudo-code (\cref{alg}) is presented to more vividly demonstrate the overall pipeline.

\section{Experiments}

\begin{table*}[]
\centering
\begin{tabular}{llcccccc}
\hline
\multicolumn{1}{c}{Models}                       & \multicolumn{1}{c}{Methods} & E5M8             & E5M7             & E5M6             & E5M5             & E5M4             & E5M3             \\ \hline
\multirow{4}{*}{LLaMA2-7B}   & Before Fine-Tuning          & 57.64\%          & 57.65\%          & 57.64\%          & 57.42\%          & 57.79\%          & 56.99\%          \\
                             & FP16 Fine-Tuning            & 58.24\%          & 58.27\%          & 58.14\%          & 57.96\%          & 57.82\%          & 56.46\%          \\
                             & Fixed Precision Fine-Tuning & 58.31\%          & 58.32\%          & 58.27\%          & 58.20\%          & 58.14\%          & 57.09\%          \\
                             & \textbf{Ours}               & \textbf{59.15\%} & \textbf{59.09\%} & \textbf{58.93\%} & \textbf{58.75\%} & \textbf{58.70\%} & \textbf{57.72\%} \\ \hline
\multirow{4}{*}{LLaMA2-13B}  & Before Fine-Tuning          & 60.87\%          & 60.75\%          & 60.76\%          & 60.87\%          & 60.40\%          & 60.41\%          \\
                             & FP16 Fine-Tuning            & 61.43\%          & 61.42\%          & 61.43\%          & 61.13\%          & 60.84\%          & 60.01\%          \\
                             & Fixed Precision Fine-Tuning & 61.54\%          & 61.43\%          & 61.50\%          & 61.17\%          & 61.23\%          & 60.63\%          \\
                             & \textbf{Ours}               & \textbf{62.33\%} & \textbf{62.05\%} & \textbf{62.11\%} & \textbf{61.83\%} & \textbf{61.89\%} & \textbf{61.10\%} \\ \hline
\multirow{4}{*}{LLaMA3-8B}   & Before Fine-Tuning          & 62.44\%          & 62.35\%          & 62.49\%          & 62.31\%          & 62.01\%          & 59.81\%          \\
                             & FP16 Fine-Tuning            & 63.42\%          & 63.50\%          & 63.44\%          & 63.29\%          & 61.23\%          & 59.62\%          \\
                             & Fixed Precision Fine-Tuning & 63.89\%          & 63.64\%          & 63.55\%          & 63.34\%          & 62.30\%          & 59.84\%          \\
                             & \textbf{Ours}               & \textbf{64.09\%} & \textbf{64.12\%} & \textbf{63.99\%} & \textbf{63.84\%} & \textbf{63.34\%} & \textbf{60.67\%} \\ \hline
\multirow{4}{*}{LLaMA3.2-3B} & Before Fine-Tuning          & 57.21\%          & 57.10\%          & 57.39\%          & 57.29\%          & 56.68\%          & 54.52\%          \\
                             & FP16 Fine-Tuning            & 57.80\%          & 57.75\%          & 57.92\%          & 57.74\%          & 56.78\%          & 54.57\%          \\
                             & Fixed Precision Fine-Tuning & 57.89\%          & 58.08\%          & 58.14\%          & 57.85\%          & 57.00\%          & 54.78\%          \\
                             & \textbf{Ours}               & \textbf{58.48\%} & \textbf{58.46\%} & \textbf{58.64\%} & \textbf{58.21\%} & \textbf{57.82\%} & \textbf{56.03\%} \\ \hline
\multirow{4}{*}{Qwen3-8B}    & Before Fine-Tuning          & 64.45\%          & 64.38\%          & 64.72\%          & 63.29\%          & 63.13\%          & 58.65\%          \\
                             & FP16 Fine-Tuning            & 65.82\%          & 65.62\%          & 65.80\%          & 64.95\%          & 63.62\%          & 58.60\%          \\
                             & Fixed Precision Fine-Tuning & 65.93\%          & 66.10\%          & 65.91\%          & 65.14\%          & 64.82\%          & 61.05\%          \\
                             & \textbf{Ours}               & \textbf{66.74\%} & \textbf{66.64\%} & \textbf{66.58\%} & \textbf{66.19\%} & \textbf{66.13\%} & \textbf{61.51\%} \\ \hline
\end{tabular}
\label{generalization}
\caption{Zero-shot results of LLaMA2-7B, LLaMA2-13B, LLaMA3-8B, LLaMA3.2-3B, Qwen3-8B. We report average accuracies of all zero-shot tasks, i.e., Arc-Challenge, Arc-Easy, BoolQ, HellaSwag, MATHQA, OpenBookQA, PIQA, WinoGrande.}
\label{overall}
\end{table*}

In this section, we present a comprehensive evaluation of the proposed OTARo method. We report the experimental setup, i.e., models, datasets, methods and implementation details, and show the overall results. In addition, we perform ablation studies, and calculate the memory reduction and speedup of SEFP quantization. 
\begin{figure}[!t]
    \centering
    \includegraphics[width=0.75\linewidth, trim={10pt 10pt 0pt 10pt}, clip]{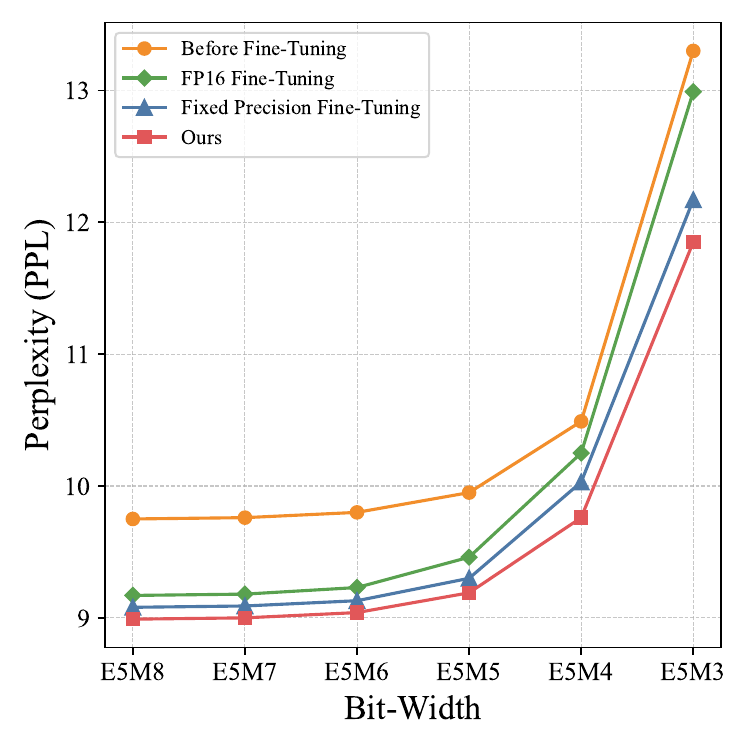}
    \caption{Task-specific fine-tuning results of LLaMA3.2-1B.}
    \label{fig-wikit}
\end{figure}

\begin{figure*}[!t]
    \centering
    \subfloat[strategies comparison]{
    		\includegraphics[scale=0.453, trim={20pt 21pt 0pt 0pt}, clip]{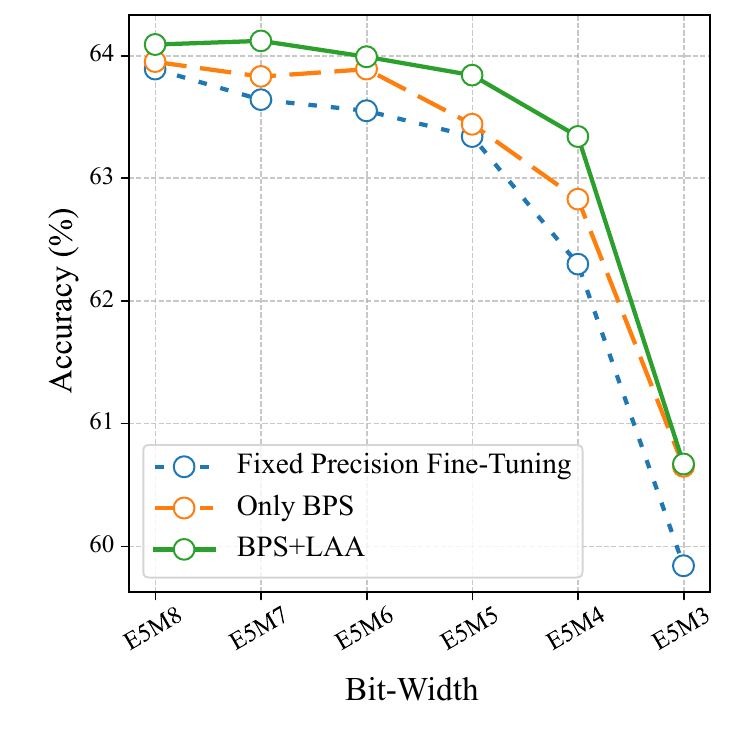}}
    \subfloat[exploration coefficient $\lambda$]{
    		\includegraphics[scale=0.45, trim={0pt 15pt 0pt 0pt}, clip]{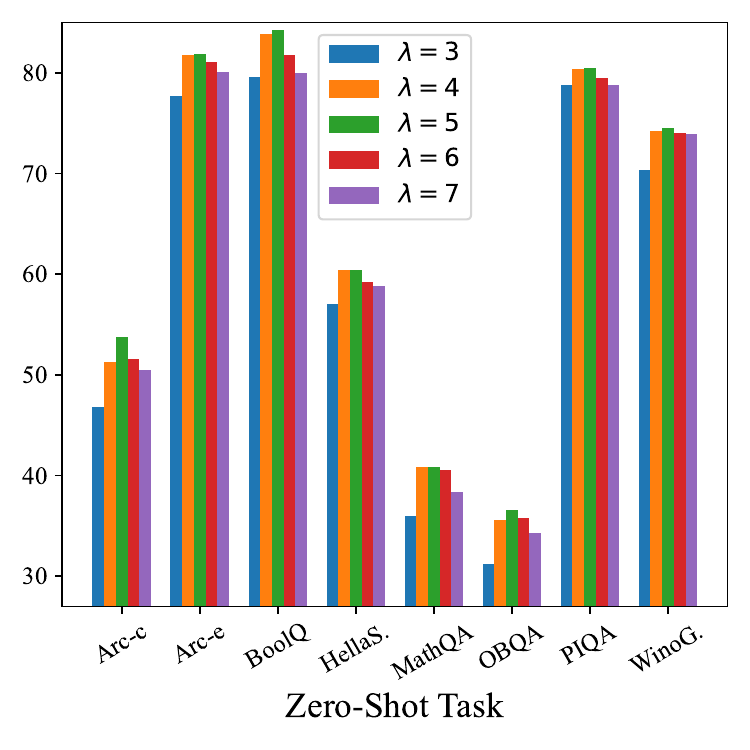}}
    \subfloat[delay step $N$]{
    		\includegraphics[scale=0.45, trim={10pt 15pt 0pt 0pt}, clip]{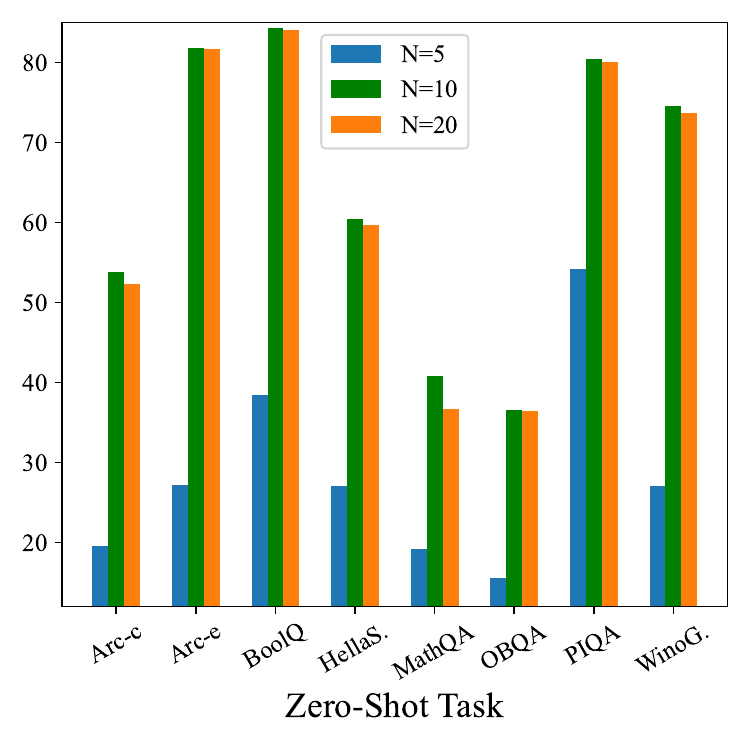}}

    \caption{Ablation results. We show the ablation results for strategies, exploration coefficient $\lambda$, delay step $N$. For strategies comparison, we report average accuracies of all zero-shot tasks, and for the others, we report E5M8 accuracies in each task.}
    \label{fig-ablation}
\end{figure*}

\subsection{Setup}
\subsubsection{Models} In the current development of LLMs, LLaMA \cite{touvron2023llama2,grattafiori2024llama} and Qwen\cite{yang2025qwen3} families have gained popularity due to strong performance and broad community support. Accordingly, in zero-shot experiments, we use LLaMA2-7B, LLaMA2-13B, LLaMA3-8B, LLaMA3.2-3B and Qwen3-8B, and in task-specific fine-tuning experiments, we use LLaMA3.2-1B.

\subsubsection{Datasets} For zero-shot experiments, we adopt the Alpaca \cite{alpaca} dataset.
Each model is fine-tuned on Alpaca, and evaluated on 
a suite of benchmarks spanning different reasoning and understanding skills:
ARC-Easy, ARC-Challenge \cite{clark2018think}, BoolQ \cite{clark2019boolq}, HellaSwag \cite{zellers2019hellaswag}, MATHQA \cite{amini2019mathqa}, OpenBookQA \cite{mihaylov2018bookqa}, PIQA \cite{bisk2020piqa}, WinoGrande \cite{sakaguchi2021winogrande}. 
For task-specific fine-tuning experiments, we adopt the WikiText2 \cite{merity2016pointer} dataset. The model is fine-tuned and evaluated using its train and test set, respectively.

\subsubsection{Methods} We evaluate pre-trained models to show the optimizations from fine-tuning. FP16 fine-tuning and fixed precision fine-tuning are also included as baselines to highlight the bit-width robustness achieved by OTARo. 
In fixed precision fine-tuning, the backpropagation-based weight update mechanism aligns with that employed in OTARo. Fixed precision fine-tuning make the models adapt to quantization errors from each fixed bit-width, while multiplying the total fine-tuning time.

\subsubsection{Implementation Details} In the experiments, we set the learning rate to 1e-5, and use SGD optimizer. SEFP group size is set to 64, and the bit-widths include \{E5M8, E5M7, E5M6, E5M5, E5M4, E5M3\}. In the BPS strategy, exploration coefficient $\lambda$ is set to 5, and in the LAA strategy, delayed updates are performed with a delay step $N$=10. For zero-shot experiments, each model is fine-tuned in a single epoch to avoid excessive adaptation, and evaluated with accuracy (\%). For task-specific fine-tuning experiments, the model is fine-tuned in 20 epochs for a well-converged state, and evaluated with perplexity, i.e. PPL. In ablation studies, strategies comparison, $\lambda$, and $N$ are included. All experiments are conducted on NVIDIA A6000 GPUs.

\subsection{Zero-Shot Results}
We report overall results of zero-shot experiments in \cref{overall}, where OTARo consistently achieves the highest average accuracy across all bit-widths (E5M8 to E5M3) for all models. 

For challenging low-bit settings (E5M4, E5M3), OTARo also obtains high accuracies and outperforms the baselines. For example, in LLaMA3-8B, OTARo achieves 63.34\% at E5M4 and 60.67\% at E5M3, compared with 62.30\% and 59.84\% from fixed precision fine-tuning, and 61.23\% and 59.62\% from FP16 fine-tuning. And, in Qwen3-8B, OTARo reaches 66.13\%, 61.51\% respectively at E5M4, E5M3, exceeding both fixed precision fine-tuning results (64.82\%, 61.05\%) and FP16 fine-tuning results (63.62\%, 58.60\%). 

These experimental results highlight the ability of OTARo to maintain robust performance in zero-shot tasks.
Detailed results are provided in Appendix C.

\subsection{Task-Specific Fine-Tuning Performances}
As shown in \cref{fig-wikit}, in the task-specific fine-tuning task, OTARo outperforms all baselines, achieving the lowest perplexity in all bit-widths. Detailed data is in Appendix D, where we can see, compared to fixed precision fine-tuning, OTARo obtains an average reduction of 0.16 PPL, and at lower bit-widths, the benefits are more pronounced, with reductions of 0.27 and 0.32 PPL respectively at E5M4 and E5M3. The experimental results confirm the effectiveness of OTARo in task-specific fine-tuning.

\subsection{Ablation Studies}

We perform ablation studies to evaluate the impacts of the BPS and LAA strategies. BPS alone can achieve robustness by exploring multiple bit-widths rather than relying on a fixed one. However, BPS-only fine-tuned models still suffer from low-precision gradient oscillations. By incorporating LAA, the model obtains further performance improvements.

Moreover, we ablate exploration coefficient $\lambda$ in BPS. 
A value too large causes an under-use of high precisions, whereas a value too small leads to an insufficient exploration of low precisions. 
Experiments with $\lambda \in$ \{3, 4, 5, 6, 7\} confirm that $\lambda$=5 offers the best balance.

We also explore different values of delay step $N$ in LAA. $N$=10 strikes a balance between smoothing gradient oscillations and ensuring sufficient updates, achieving the best performance compared to $N$=5 and $N$=20.

Ablation results are shown in \cref{fig-ablation}.

\subsection{Memory and Speed}
We benchmark the memory consumption (including storage spaces for weights and KV cache) and decoding throughput of LLaMA3-8B respectively under FP16 and SEFP data formats, suppose an input of 2000 tokens, to show the efficiency gains from SEFP quantization. As shown in \cref{tab-ms}, from FP16 to SEFP, the memory consumption is reduced by 69\%, and the speedup of decoding throughput reaches $\times$2.45.
These results highlight the practical advantages of SEFP in enabling light-weight and high-throughput on-device LLMs.

\begin{table}[]
\centering
\begin{tabular}{ccc}
\hline
Precisions & Mem. (GB) & Dec. Thpt. (token/s) \\ \hline
FP16      & 15.20       & 39.00                       \\
SEFP-E5M4      & 4.77 ($69\%\downarrow$)        & 95.65 ($\times 2.45$)                       \\ \hline
\end{tabular}
\caption{The memory consumption (Mem.) and decoding throughput (Dec. Thpt.) of FP16 and SEFP LLaMA-8B (E5M4 taken as an example for SEFP).}
\label{tab-ms}
\end{table}


\section{Conclusion}
In this paper, we have proposed OTARo, a novel fine-tuning method that enables robust, flexible, and hardware-friendly LLMs at edge. 
OTARo can obtain one unified model to support all precisions after once fine-tuning.
It introduces SEFP quantization to switch precisions flexibly, and performs a complete end-to-end learning workflow that stabilizes the performances across precisions.
Experiments demonstrate its effectiveness in multiple LLMs and domains.
In summary, this work advances the multi-precision adaptation of on-device LLMs, and provides a crucial methodological support for the real-world edge intelligence.

\bigskip

\bibliography{aaai2026}

@article{zubiaga2024natural,
  title={Natural language processing in the era of large language models.},
  author={Zubiaga, A},
  journal={Front Artif Intell},
  year={2024},
  publisher={Frontiers}
}

@article{qu2025mobile,
  title={Mobile edge intelligence for large language models: A contemporary survey},
  author={Qu, Guanqiao and Chen, Qiyuan and Wei, Wei and Lin, Zheng and Chen, Xianhao and Huang, Kaibin},
  journal={IEEE Communications Surveys \& Tutorials},
  year={2025},
  publisher={IEEE}
}

@inproceedings{tang2025scaling,
  title={Scaling On-Device GPU Inference for Large Generative Models},
  author={Tang, Jiuqiang and Sorokin, Raman and Ignasheva, Ekaterina and Jensen, Grant and Chen, Lin and Lee, Juhyun and Kulik, Andrei and Grundman, Matthias},
  booktitle={Proceedings of the Computer Vision and Pattern Recognition Conference},
  pages={6355--6364},
  year={2025}
}

@article{frantar2022gptq,
  title={Gptq: Accurate post-training quantization for generative pre-trained transformers},
  author={Frantar, Elias and Ashkboos, Saleh and Hoefler, Torsten and Alistarh, Dan},
  journal={arXiv preprint arXiv:2210.17323},
  year={2022}
}

@article{lin2024awq,
  title={Awq: Activation-aware weight quantization for on-device llm compression and acceleration},
  author={Lin, Ji and Tang, Jiaming and Tang, Haotian and Yang, Shang and Chen, Wei-Ming and Wang, Wei-Chen and Xiao, Guangxuan and Dang, Xingyu and Gan, Chuang and Han, Song},
  journal={Proceedings of Machine Learning and Systems},
  volume={6},
  pages={87--100},
  year={2024}
}

@article{liu2023llm,
  title={Llm-qat: Data-free quantization aware training for large language models},
  author={Liu, Zechun and Oguz, Barlas and Zhao, Changsheng and Chang, Ernie and Stock, Pierre and Mehdad, Yashar and Shi, Yangyang and Krishnamoorthi, Raghuraman and Chandra, Vikas},
  journal={arXiv preprint arXiv:2305.17888},
  year={2023}
}

@article{chen2024efficientqat,
  title={Efficientqat: Efficient quantization-aware training for large language models},
  author={Chen, Mengzhao and Shao, Wenqi and Xu, Peng and Wang, Jiahao and Gao, Peng and Zhang, Kaipeng and Luo, Ping},
  journal={arXiv preprint arXiv:2407.11062},
  year={2024}
}

@article{ashkboos2024quarot,
  title={Quarot: Outlier-free 4-bit inference in rotated llms},
  author={Ashkboos, Saleh and Mohtashami, Amirkeivan and Croci, Maximilian L and Li, Bo and Cameron, Pashmina and Jaggi, Martin and Alistarh, Dan and Hoefler, Torsten and Hensman, James},
  journal={Advances in Neural Information Processing Systems},
  volume={37},
  pages={100213--100240},
  year={2024}
}

@article{liu2024spinquant,
  title={Spinquant: Llm quantization with learned rotations},
  author={Liu, Zechun and Zhao, Changsheng and Fedorov, Igor and Soran, Bilge and Choudhary, Dhruv and Krishnamoorthi, Raghuraman and Chandra, Vikas and Tian, Yuandong and Blankevoort, Tijmen},
  journal={arXiv preprint arXiv:2405.16406},
  year={2024}
}

@article{wei2024advances,
  title={Advances in the neural network quantization: A comprehensive review},
  author={Wei, Lu and Ma, Zhong and Yang, Chaojie and Yao, Qin},
  journal={Applied Sciences},
  volume={14},
  number={17},
  pages={7445},
  year={2024},
  publisher={MDPI}
}

@article{gao2024precision,
  title={Precision-Aware Iterative Algorithms Based on Group-Shared Exponents of Floating-Point Numbers},
  author={Gao, Jianhua and Shen, Jiayuan and Zhang, Yuxiang and Ji, Weixing and Huang, Hua},
  journal={arXiv preprint arXiv:2411.04686},
  year={2024}
}

@article{qiao2025tellme,
  title={TeLLMe: An Energy-Efficient Ternary LLM Accelerator for Prefilling and Decoding on Edge FPGAs},
  author={Qiao, Ye and Chen, Zhiheng and Zhang, Yifan and Wang, Yian and Huang, Sitao},
  journal={arXiv preprint arXiv:2504.16266},
  year={2025}
}

@article{zhou2025gsq,
  title={GSQ-Tuning: Group-Shared Exponents Integer in Fully Quantized Training for LLMs On-Device Fine-tuning},
  author={Zhou, Sifan and Wang, Shuo and Yuan, Zhihang and Shi, Mingjia and Shang, Yuzhang and Yang, Dawei},
  journal={arXiv preprint arXiv:2502.12913},
  year={2025}
}

@article{lu2025fine,
  title={Fine-tuning large language models for domain adaptation: Exploration of training strategies, scaling, model merging and synergistic capabilities},
  author={Lu, Wei and Luu, Rachel K and Buehler, Markus J},
  journal={npj Computational Materials},
  volume={11},
  number={1},
  pages={84},
  year={2025},
  publisher={Nature Publishing Group UK London}
}

@article{yang2024unveiling,
  title={Unveiling the generalization power of fine-tuned large language models},
  author={Yang, Haoran and Zhang, Yumeng and Xu, Jiaqi and Lu, Hongyuan and Heng, Pheng Ann and Lam, Wai},
  journal={arXiv preprint arXiv:2403.09162},
  year={2024}
}

@article{manduchi2024challenges,
  title={On the challenges and opportunities in generative ai},
  author={Manduchi, Laura and Pandey, Kushagra and Meister, Clara and Bamler, Robert and Cotterell, Ryan and D{\"a}ubener, Sina and Fellenz, Sophie and Fischer, Asja and G{\"a}rtner, Thomas and Kirchler, Matthias and others},
  journal={arXiv preprint arXiv:2403.00025},
  year={2024}
}

@article{kalliojarvi2002roundoff,
  title={Roundoff errors in block-floating-point systems},
  author={Kalliojarvi, Kari and Astola, Jaakko},
  journal={IEEE transactions on signal processing},
  volume={44},
  number={4},
  pages={783--790},
  year={2002},
  publisher={IEEE}
}

@article{achiam2023gpt,
  title={Gpt-4 technical report},
  author={Achiam, Josh and Adler, Steven and Agarwal, Sandhini and Ahmad, Lama and Akkaya, Ilge and Aleman, Florencia Leoni and Almeida, Diogo and Altenschmidt, Janko and Altman, Sam and Anadkat, Shyamal and others},
  journal={arXiv preprint arXiv:2303.08774},
  year={2023}
}

@article{touvron2023llama2,
  title={Llama 2: Open foundation and fine-tuned chat models},
  author={Touvron, Hugo and Martin, Louis and Stone, Kevin and Albert, Peter and Almahairi, Amjad and Babaei, Yasmine and Bashlykov, Nikolay and Batra, Soumya and Bhargava, Prajjwal and Bhosale, Shruti and others},
  journal={arXiv preprint arXiv:2307.09288},
  year={2023}
}

@article{grattafiori2024llama,
  title={The llama 3 herd of models},
  author={Grattafiori, Aaron and Dubey, Abhimanyu and Jauhri, Abhinav and Pandey, Abhinav and Kadian, Abhishek and Al-Dahle, Ahmad and Letman, Aiesha and Mathur, Akhil and Schelten, Alan and Vaughan, Alex and others},
  journal={arXiv preprint arXiv:2407.21783},
  year={2024}
}

@article{yang2025qwen3,
  title={Qwen3 technical report},
  author={Yang, An and Li, Anfeng and Yang, Baosong and Zhang, Beichen and Hui, Binyuan and Zheng, Bo and Yu, Bowen and Gao, Chang and Huang, Chengen and Lv, Chenxu and others},
  journal={arXiv preprint arXiv:2505.09388},
  year={2025}
}

@article{yin2019understanding,
  title={Understanding straight-through estimator in training activation quantized neural nets},
  author={Yin, Penghang and Lyu, Jiancheng and Zhang, Shuai and Osher, Stanley and Qi, Yingyong and Xin, Jack},
  journal={arXiv preprint arXiv:1903.05662},
  year={2019}
}

@misc{alpaca,
  author = {Rohan Taori and Ishaan Gulrajani and Tianyi Zhang and Yann Dubois and Xuechen Li and Carlos Guestrin and Percy Liang and Tatsunori B. Hashimoto },
  title = {Stanford Alpaca: An Instruction-following LLaMA model},
  year = {2023},
  publisher = {GitHub},
  journal = {GitHub repository},
  howpublished = {\url{https://github.com/tatsu-lab/stanford_alpaca}},
}

@article{merity2016pointer,
  title={Pointer sentinel mixture models},
  author={Merity, Stephen and Xiong, Caiming and Bradbury, James and Socher, Richard},
  journal={arXiv preprint arXiv:1609.07843},
  year={2016}
}

@article{clark2019boolq,
  title={Boolq: Exploring the surprising difficulty of natural yes/no questions},
  author={Clark, Christopher and Lee, Kenton and Chang, Ming-Wei and Kwiatkowski, Tom and Collins, Michael and Toutanova, Kristina},
  journal={arXiv preprint arXiv:1905.10044},
  year={2019}
}

@inproceedings{bisk2020piqa,
  title={Piqa: Reasoning about physical commonsense in natural language},
  author={Bisk, Yonatan and Zellers, Rowan and Gao, Jianfeng and Choi, Yejin and others},
  booktitle={Proceedings of the AAAI conference on artificial intelligence},
  volume={34},
  number={05},
  pages={7432--7439},
  year={2020}
}

@article{zellers2019hellaswag,
  title={Hellaswag: Can a machine really finish your sentence?},
  author={Zellers, Rowan and Holtzman, Ari and Bisk, Yonatan and Farhadi, Ali and Choi, Yejin},
  journal={arXiv preprint arXiv:1905.07830},
  year={2019}
}

@article{mihaylov2018bookqa,
  title={Can a suit of armor conduct electricity? a new dataset for open book question answering},
  author={Mihaylov, Todor and Clark, Peter and Khot, Tushar and Sabharwal, Ashish},
  journal={arXiv preprint arXiv:1809.02789},
  year={2018}
}

@article{sakaguchi2021winogrande,
  title={Winogrande: An adversarial winograd schema challenge at scale},
  author={Sakaguchi, Keisuke and Bras, Ronan Le and Bhagavatula, Chandra and Choi, Yejin},
  journal={Communications of the ACM},
  volume={64},
  number={9},
  pages={99--106},
  year={2021},
  publisher={ACM New York, NY, USA}
}

@article{clark2018think,
  title={Think you have solved question answering? try arc, the ai2 reasoning challenge},
  author={Clark, Peter and Cowhey, Isaac and Etzioni, Oren and Khot, Tushar and Sabharwal, Ashish and Schoenick, Carissa and Tafjord, Oyvind},
  journal={arXiv preprint arXiv:1803.05457},
  year={2018}
}

@article{amini2019mathqa,
  title={Mathqa: Towards interpretable math word problem solving with operation-based formalisms},
  author={Amini, Aida and Gabriel, Saadia and Lin, Peter and Koncel-Kedziorski, Rik and Choi, Yejin and Hajishirzi, Hannaneh},
  journal={arXiv preprint arXiv:1905.13319},
  year={2019}
}

@article{yang2024post,
  title={Post-training quantization for re-parameterization via coarse \& fine weight splitting},
  author={Yang, Dawei and He, Ning and Hu, Xing and Yuan, Zhihang and Yu, Jiangyong and Xu, Chen and Jiang, Zhe},
  journal={Journal of Systems Architecture},
  volume={147},
  pages={103065},
  year={2024},
  publisher={Elsevier}
}

@article{park2024any,
  title={Any-precision llm: Low-cost deployment of multiple, different-sized llms},
  author={Park, Yeonhong and Hyun, Jake and Cho, SangLyul and Sim, Bonggeun and Lee, Jae W},
  journal={arXiv preprint arXiv:2402.10517},
  year={2024}
}

@misc{zhao2025quark,
      title={QUARK: Quantization-Enabled Circuit Sharing for Transformer Acceleration by Exploiting Common Patterns in Nonlinear Operations}, 
      author={Zhixiong Zhao and Haomin Li and Fangxin Liu and Yuncheng Lu and Zongwu Wang and Tao Yang and Li Jiang and Haibing Guan},
      year={2025},
      eprint={2511.06767},
      archivePrefix={arXiv},
      primaryClass={cs.LG},
      url={https://arxiv.org/abs/2511.06767}, 
}

\clearpage
\appendix

\section{A. Visual $\epsilon(\omega_0)$ of Different Mantissa Bit-Widths}
\begin{figure}[!t]
    \centering
    \includegraphics[width=1\linewidth]{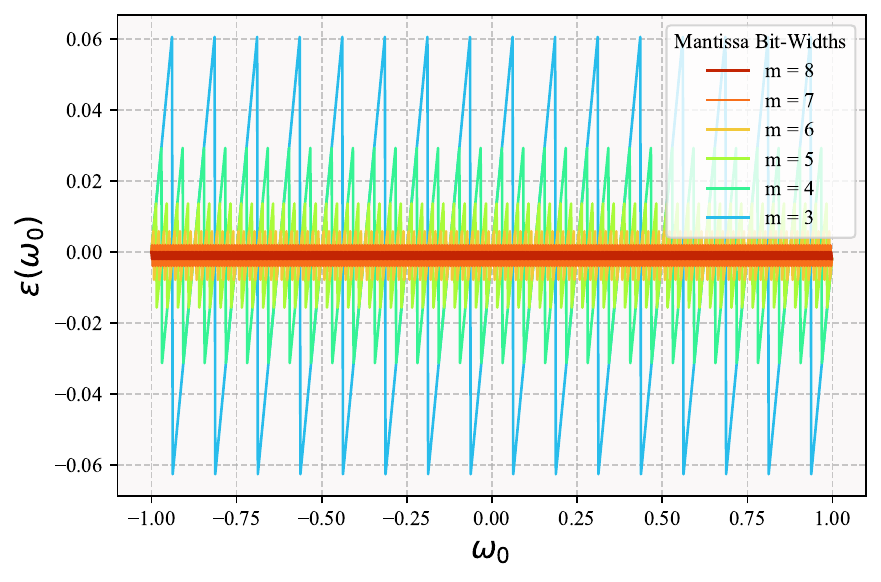}
    \caption{Visualization of $\epsilon(\omega_0)$ function under the mantissa bit-width $m=\{8,7,6,5,4,3\}$.}
    \label{fig-epsilon}
\end{figure}

In this section, we illustrate the $\epsilon(\omega_0)$ of different mantissa bit-widths ($m=\{8,7,6,5,4,3\}$), as shown in \cref{fig-epsilon}.

\section{B. High-Dimensional Linear Mapping}
To further investigate the gradient oscillations introduced by low bit-width quantization, we adopt a high-dimensional linear mapping approach. Specifically, we model the relationship between the gradient under SEFP fine-tuning, $\nabla_{\text{sefp}}$, and the gradient under FP16 fine-tuning, $\nabla_{\text{fp}}$, as the following linear relationship:
\begin{equation}
    \nabla_{\text{sefp}} = X \cdot \nabla_{\text{fp}} + Y,
\end{equation}
where $X$ denotes a linear mapping matrix, and $Y$ represents the residual term that captures additional perturbations introduced by quantization.

To solve for $X$ and $Y$ in the above formulation, we collect the gradient samples from $N$ consecutive batches, which are denoted as:
\begin{equation}
    G_{\text{fp}} = \begin{bmatrix} \nabla_{\text{fp},1} \\ \nabla_{\text{fp},2} \\ \vdots \\ \nabla_{\text{fp},N} \end{bmatrix} \in \mathbb{R}^{N \times d}, \quad
    G = \begin{bmatrix} \nabla_{\text{sefp},1} \\ \nabla_{\text{sefp},2} \\ \vdots \\ \nabla_{\text{sefp},N} \end{bmatrix} \in \mathbb{R}^{N \times d},
\end{equation}
where $d$ denotes the dimensionality of the model parameters.

According to the principle of Least Squares
Method (LSM), the optimal solution for $X$ minimizes the reconstruction error $\| G - G_{\text{fp}} X \|_F^2$, and its analytical solution is given by:
\begin{equation}
    X = (G_{\text{fp}}^\top G_{\text{fp}})^{-1} G_{\text{fp}}^\top G,
\end{equation}
where $(\cdot)^\top$ denotes matrix transpose, and $\|\cdot\|_F$ represents the Frobenius norm.

After obtaining $X$, the residual perturbation corresponding to each batch is calculated as:
\begin{equation}
    Y = G - G_{\text{fp}} X,
\end{equation}
that is,
\begin{equation}
    Y_i = \nabla_{\text{sefp},i} - X \cdot \nabla_{\text{fp},i}, \quad i=1,\ldots,N.
\end{equation}

According to this formulation, $Y$ completely captures the residual stochastic perturbations in the gradients under SEFP quantization that cannot be explained by linear mapping from $\nabla_{\text{fp}}$. This enables a more accurate characterization of the non-linear errors introduced by low bit-width quantization.

Compared to the conventional direct differencing method (i.e., $\nabla_{\text{sefp}} - \nabla_{\text{fp}}$), this approach introduces a fitted $X$ term, which effectively eliminates the linear scaling effect caused by the variation in gradient magnitude across batches. Therefore, $Y$ is able to more purely reflect the stochastic and non-stationary nature of the SEFP-induced perturbations.

\section{C. Detailed Zero-Shot Results}
In zero-shot experiments, we include 5 popular LLMs, i.e., LLaMA2-7B, LLaMA2-13B, LLaMA3-8B, LLaMA3.2-3B, Qwen3-8B. Each model is fine-tuned with Alpaca dataset and tested with ARC-Easy, ARC-Challenge, BoolQ, HellaSwag, MATHQA, OpenBookQA, PIQA, WinoGrande. The detailed zero-shot results of these LLMs are demonstrated in \cref{llama2-7b,llama2-13b,llama3-8b,LLaMa3.2-3B,qwen}.

\begin{table*}[!t]
\centering

\begin{tabular}{clcccccc}
\hline
Methods                                         & \multicolumn{1}{c}{Tasks}        & E5M8                                     & E5M7                                     & E5M6                                     & E5M5                                     & E5M4                                     & E5M3                                     \\ \hline
                                                & Arc-c                                 & 43.34\%                                  & 43.34\%                                  & 43.09\%                                  & 42.41\%                                  & 42.75\%                                  & 43.77\%                                  \\
                                                & Arc-e                                 & 76.35\%                                  & 76.39\%                                  & 76.18\%                                  & 76.47\%                                  & 76.30\%                                  & 75.25\%                                  \\
                                                & BoolQ                                 & 77.74\%                                  & 77.74\%                                  & 77.40\%                                  & 77.06\%                                  & 78.47\%                                  & 76.67\%                                  \\
                                                & HellaS.                               & 57.14\%                                  & 57.20\%                                  & 57.16\%                                  & 57.08\%                                  & 56.90\%                                  & 56.42\%                                  \\
\hline
\end{tabular}
\caption{Detailed zero-shot results of LLaMA2-7B. We report accuracies in Arc-Challenge, Arc-Easy, BoolQ, HellaSwag, MATHQA, OpenBookQA, PIQA, WinoGrande, and the average values of all zero-shot tasks.}
\label{llama2-7b}
\end{table*}
\begin{table*}[!t]
\centering

\begin{tabular}{clcccccc}
\hline
Methods                                         & \multicolumn{1}{c}{Tasks}        & E5M8                                     & E5M7                                     & E5M6                                     & E5M5                                     & E5M4                                     & E5M3                                     \\ \hline
                                                & Arc-c                                 & 48.38\%                                  & 47.95\%                                  & 47.87\%                                  & 48.46\%                                  & 47.78\%                                  & 47.10\%                                  \\
                                                & Arc-e                                 & 79.38\%                                  & 79.34\%                                  & 78.96\%                                  & 79.55\%                                  & 79.29\%                                  & 79.00\%                                  \\
                                                & BoolQ                                 & 80.58\%                                  & 80.34\%                                  & 80.40\%                                  & 80.40\%                                  & 79.51\%                                  & 81.04\%                                  \\
                                                & HellaS.                               & 60.05\%                                  & 60.06\%                                  & 60.12\%                                  & 59.98\%                                  & 59.62\%                                  & 58.84\%                                  \\
\hline
\end{tabular}

\caption{Detailed zero-shot results of LLaMA2-13B. We report accuracies in Arc-Challenge, Arc-Easy, BoolQ, HellaSwag, MATHQA, OpenBookQA, PIQA, WinoGrande, and the average values of all zero-shot tasks.}
\label{llama2-13b}

\end{table*}
\begin{table*}[!t]
\centering

\begin{tabular}{clcccccc}
\hline
Methods                                       & \multicolumn{1}{c}{Tasks}                           & E5M8                                     & E5M7                                     & E5M6                                     & E5M5                                     & E5M4                                     & E5M3                                     \\ \hline
                                             & Arc-c                                 & 50.43\%                                  & 50.34\%                                  & 50.60\%                                  & 50.26\%                                  & 49.15\%                                  & 45.48\%                                  \\
                                             & Arc-e                                 & 80.18\%                                  & 79.80\%                                  & 80.35\%                                  & 80.01\%                                  & 79.42\%                                  & 76.73\%                                  \\
                                             & BoolQ                                 & 81.22\%                                  & 80.95\%                                  & 81.25\%                                  & 80.46\%                                  & 80.89\%                                  & 79.82\%                                  \\
                                             & HellaS.                               & 60.16\%                                  & 60.13\%                                  & 60.08\%                                  & 59.99\%                                  & 59.89\%                                  & 57.97\%                                  \\
\hline
\end{tabular}

\caption{Detailed zero-shot results of LLaMA3-8B. We report accuracies in Arc-Challenge, Arc-Easy, BoolQ, HellaSwag, MATHQA, OpenBookQA, PIQA, WinoGrande, and the average values of all zero-shot tasks.}

\label{llama3-8b}

\end{table*}
\begin{table*}[t]
\centering
\begin{tabular}{clcccccc}
\hline
Methods                                       & \multicolumn{1}{c}{Tasks}        & E5M8                                     & E5M7                                     & E5M6                                     & E5M5                                     & E5M4                                     & E5M3                                     \\ \hline
                                              & Arc-c                                 & 41.89\%                                  & 42.15\%                                  & 42.32\%                                  & 42.32\%                                  & 42.66\%                                  & 37.46\%                                  \\
                                              & Arc-e                                 & 74.82\%                                  & 74.58\%                                  & 74.58\%                                  & 74.45\%                                  & 74.28\%                                  & 69.07\%                                  \\
                                              & BoolQ                                 & 73.46\%                                  & 73.18\%                                  & 73.76\%                                  & 74.04\%                                  & 70.83\%                                  & 72.87\%                                  \\
                                              & HellaS.                               & 55.31\%                                  & 55.29\%                                  & 55.36\%                                  & 55.37\%                                  & 54.55\%                                  & 52.42\%                                  \\
\hline
\end{tabular}
\caption{Detailed zero-shot results of LLaMA3.2-3B. We report accuracies in Arc-Challenge, Arc-Easy, BoolQ, HellaSwag, MATHQA, OpenBookQA, PIQA, WinoGrande, and the average values of all zero-shot tasks.}
\label{LLaMa3.2-3B}
\end{table*}
\begin{table*}[!t]
\centering

\begin{tabular}{clcccccc}
\hline
Methods                                         & \multicolumn{1}{c}{Tasks}        & E5M8                                     & E5M7                                     & E5M6                                     & E5M5                                     & E5M4                                     & E5M3                                     \\ \hline
                                                & Arc-c                                 & 52.90\%                                  & 52.47\%                                  & 53.58\%                                  & 51.11\%                                  & 53.92\%                                  & 47.70\%                                  \\
                                                & Arc-e                                 & 81.65\%                                  & 81.69\%                                  & 82.11\%                                  & 81.02\%                                  & 81.52\%                                  & 76.98\%                                  \\
                                                & BoolQ                                 & 83.00\%                                  & 83.09\%                                  & 83.82\%                                  & 80.24\%                                  & 80.15\%                                  & 77.86\%                                  \\
                                                & HellaS.                               & 58.91\%                                  & 58.83\%                                  & 59.05\%                                  & 57.88\%                                  & 56.91\%                                  & 52.37\%                                  \\
\hline
\end{tabular}

\caption{Detailed zero-shot results of Qwen3-8B. We report accuracies in Arc-Challenge, Arc-Easy, BoolQ, HellaSwag, MATHQA, OpenBookQA, PIQA, WinoGrande, and the average values of all zero-shot tasks.}
\label{qwen}

\end{table*}

\section{D. Detailed Results of Task-Specific Fine-Tuning}
In task-specific fine-tuning experiments, we fine-tune and evaluate LLaMA3.2-1B using WikiText2 train set and test set, respectively, and report the detailed results before/after fine-tuning in \cref{tab-B}.

\begin{table*}[hbtp]
\centering
\begin{tabular}{ccccccccc}
\hline
Methods                   & E5M8 & E5M7 & E5M6 & E5M5 & E5M4  & E5M3  & AVG.  & STD.     \\ \hline
Before Fine-Tuning          & 9.75 & 9.76 & 9.80 & 9.95 & 10.49 & 13.30 & 10.51 & 1.40 \\
FP16 Fine-Tuning            & 9.17 & 9.18 & 9.23 & 9.46 & 10.25 & 12.99 & 10.05 & 1.50 \\
Fixed Precision Fine-Tuning & 9.08 & 9.09 & 9.13 & 9.30 & 10.03 & 12.17 & 9.80  & 1.26 \\
\textbf{Ours}                        & \textbf{8.99} & \textbf{9.00} & \textbf{9.04} & \textbf{9.19} & \textbf{9.76}  & \textbf{11.85} & \textbf{9.64}  & \textbf{1.12} \\ \hline
\end{tabular}
\caption{Detailed results of LLaMA3.2-1B in task-specific fine-tuning experiments. We report the perplexity (PPL) values in WikiText2 test set, and the average, standard deviation values of all bit-widths.}
\label{tab-B}
\end{table*}

\section{}

\end{document}